%% file: 00-main.tex
\documentclass[letterpaper, 10 pt, conference]{ieeeconf}  

\IEEEoverridecommandlockouts                              

\usepackage[dvipsnames, table]{xcolor}
\usepackage{graphicx}
\usepackage{siunitx}
\DeclareSIUnit{\nothing}{\relax}
\usepackage{booktabs}
\usepackage{hyperref}
\hypersetup{%
    pdfborder = {0 0 0}
}
\DeclareSIUnit{\nothing}{\relax}
\usepackage[super]{nth} 

\definecolor{somegray}{rgb}{0.5, 0.5, 0.5}
\newcommand{\darkgrayed}[1]{\textcolor{somegray}{#1}}
\makeatletter
\newcommand*\titleheader[1]{\gdef\@titleheader{#1}}
\AtBeginDocument{%
  \let\st@red@title\@title
  \def\@title{%
    \vskip-2.0em
    \bgroup\normalfont\large\centering\@titleheader\par\egroup
    \vskip0.0em\st@red@title}
}

\makeatother

\titleheader{\darkgrayed{This paper has been accepted for publication in the IEEE ICRA 2023 conference\\\copyright 2023 IEEE.}}

\title{\LARGE \bf
Ultra-low Power Deep Learning-based Monocular Relative Localization Onboard Nano-quadrotors
}

\author{S. Bonato$^{1}$, S. C. Lambertenghi$^{1}$, E. Cereda$^{1}$, A. Giusti$^{1}$, and D. Palossi$^{12}$
\thanks{This work was partially supported by the Secure Systems Research Center (SSRC) of the UAE Technology Innovation Institute (TII) and the Swiss National Science Foundation (SNSF) through the NCCR Robotics.}
\thanks{$^{1}$S. Bonato, S. C. Lambertenghi, E. Cereda, A. Giusti, and D. Palossi are with the Dalle Molle Institute for Artificial Intelligence, USI and SUPSI, Lugano, 6962, Switzerland
        {\tt\small name.surname@idsia.ch}}%
\thanks{$^{2}$D. Palossi is also with the Integrated Systems Laboratory, ETH Z\"urich, Z\"urich, 8092, Switzerland
        {\tt\small dpalossi@iis.ee.ethz.ch}}%
}

\begin{document}
\maketitle

\begin{abstract}
Precise relative localization is a crucial functional block for swarm robotics.
This work presents a novel autonomous end-to-end system that addresses the monocular relative localization, through deep neural networks (DNNs), of two peer nano-drones, i.e., sub-\SI{40}{\gram} of weight and sub-\SI{100}{\milli\watt} processing power.
To cope with the ultra-constrained nano-drone platform, we propose a vertically-integrated framework, from the dataset collection to the final in-field deployment, including dataset augmentation, quantization, and system optimizations.
Experimental results show that our DNN can precisely localize a \SI{10}{\centi\meter}-size target nano-drone by employing only low-resolution monochrome images, up to $\sim$\SI{2}{\meter} distance.
On a disjoint testing dataset our model yields a mean $R^2$ score of 0.42 and a root mean square error of \SI{18}{\centi\meter}, which results in a mean in-field prediction error of \SI{15}{\centi\meter} and in a closed-loop control error of \SI{17}{\centi\meter}, over a $\sim$\SI{60}{\second}-flight test.
Ultimately, the proposed system improves the State-of-the-Art by showing long-endurance tracking performance (up to \SI{2}{\minute} continuous tracking), generalization capabilities being deployed in a never-seen-before environment, and requiring a minimal power consumption of \SI{95}{\milli\watt} for an onboard real-time inference-rate of \SI{48}{\hertz}.
\end{abstract}

\section*{Supplementary video material}
In-field tests: \url{https://youtu.be/pUGL1qu3Z1k}.

\input{01-introduction}
\input{02-related_work}
\input{03-implementation}
\input{04-results}
\input{05-conclusion}

\bibliographystyle{./IEEEtran}
\bibliography{biblio}

\end{document}

%% file: 01-introduction.tex
\section{Introduction} \label{sec:intro}

Nano-class size quadrotors are an emerging robot platform with weights below \SI{40}{\gram} and a diameter below \SI{10}{\centi\meter}~\cite{li_self-supervised_2022,palossi2017target}.
They unlock attractive novel applications in many scenarios since they are inexpensive and can safely operate in restricted spaces and nearby humans~\cite{frontnet}.
The potential of this platform is further enhanced when many autonomous nano-drones collaborate in a group (swarm), but achieving this ambitious goal is far from trivial, needing to reliably localize each other while flying, i.e., \textit{relative localization}.
This is a significant challenge given the extreme constraints imposed by this class of robots, in terms of sensor types and quality (e.g., QVGA cameras), computational capability (a few 100s~\SI{}{\mega Ops/\second} on single-core microcontroller units), memory size (sub-\SI{10}{\mega\byte} off-chip and sub-\SI{1}{\mega\byte} on-chip memories) and power consumption (a few 100s~\SI{}{\milli\watt} compute power).

\begin{figure}[t]
  \includegraphics[width=\columnwidth]{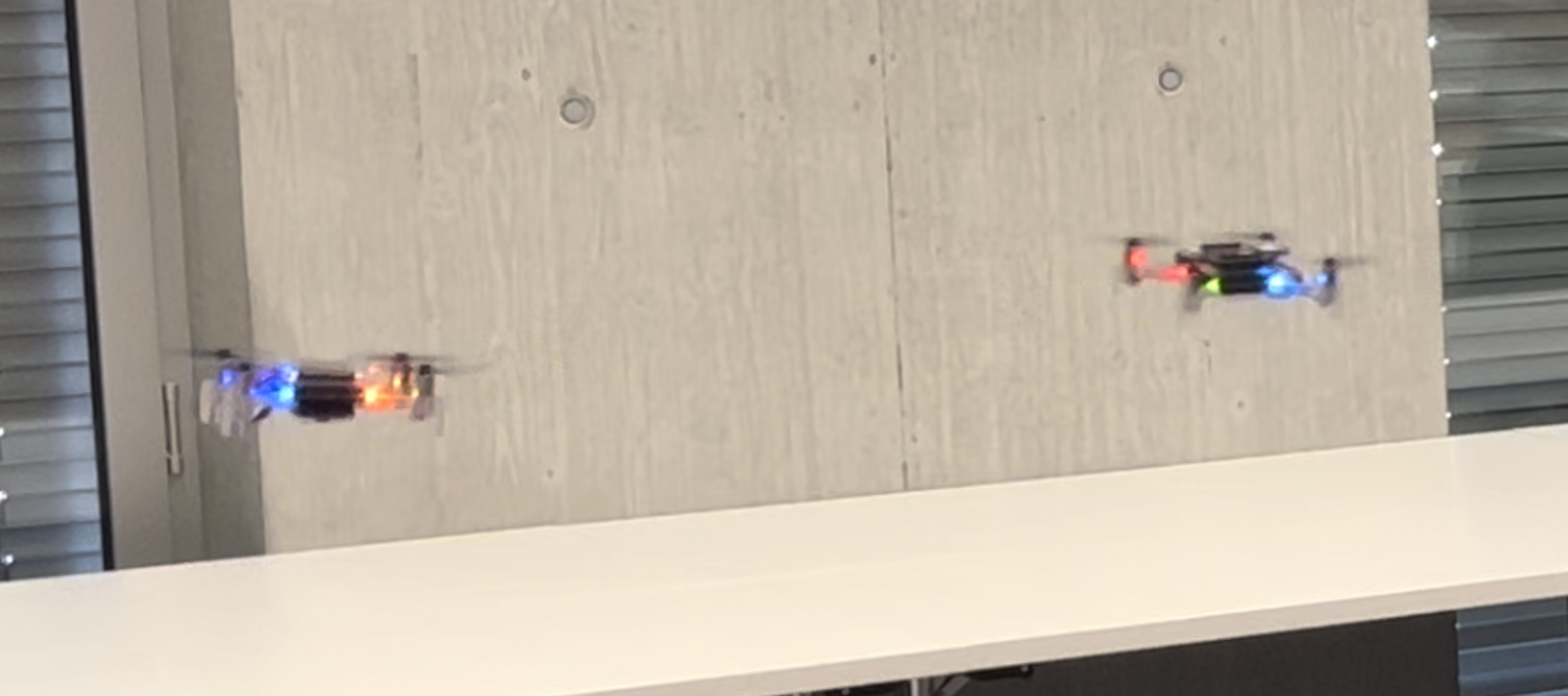}
  \caption{Our in-field system: one nano-drone visually estimates the relative pose of the other.}
  \label{fig:in_field_proto}
\end{figure}

Our work tackles this challenge by designing, implementing and thoroughly validating in the field (closed-loop) a visual relative localization approach that runs fully onboard a \SI{27}{\gram} nano-drone up to \SI{48}{frame/\second} within \SI{95}{\milli\watt} at most.
Our system, shown in Figure~\ref{fig:in_field_proto}, uses input frames from a low-resolution, low-dynamic range tiny camera; it relies on a custom deep convolutional neural network (CNN) with more than \SI{304}{\kilo\nothing} parameters, trained as a regressor to return the relative position of a peer nano-drone.
We discuss and motivate design choices spanning the whole system, including \textit{i}) acquisition setup for training and evaluation datasets; \textit{ii}) the selection of the CNN architecture and its training, quantization, and deployment on the target platform; \textit{iii}) analysis of regression performance, throughput, and computational power; \textit{iv}) integration in a control loop that tackles the task of following a peer quadrotor; \textit{v}) validation of the in-field control performance in various environments.
Despite being often overlooked with developing systems for larger robots, we highlight how some of these aspects play a fundamental role when targeting nano-quadrotors.

The main contribution of our system paper is the discussion and extensive quantitative experimental validation of the entire approach, whose video demonstration is released as supplementary material.
After reviewing related work in Section~\ref{sec:related_work}, Section~\ref{sec:implementation} presents the key aspects of the system (hardware platform, datasets, CNN's architecture/training, and the robot controller).
Before concluding the paper in Section~\ref{sec:conclusion}, Section~\ref{sec:results} outlines experimental results obtained both on static datasets and in-field, showing \textit{i}) a mean $R^2$ score of 0.42 and a root mean square error (RMSE) of \SI{18}{\centi\meter} on a disjoint testing dataset; \textit{ii}) a mean in-field prediction error of \SI{15}{\centi\meter} which maps in a closed-loop control error of \SI{17}{\centi\meter}, over a $\sim$\SI{60}{\second}-flight tests; \textit{iii}) long-endurance tracking performance, up to \SI{2}{\minute} flights; and \textit{iv}) an onboard real-time inference-rate of \SI{48}{\hertz} within \SI{95}{\milli\watt} computational power consumption.

%% file: 02-related_work.tex
\section{Related work} \label{sec:related_work}

Estimating the pose of nearby robots is key in many applications, including swarm flocking, formation flying, and obstacle avoidance during navigation, to name a few.
Many recent works in the deep learning literature deal with estimating the pose of a known object based on one or more camera images~\cite{monocular-6d-ml,carrio_drone_2018,schilling_vision-based_2021,frontnet,li_self-supervised_2022}.
In robotics, the task is typically solved using the robot's onboard sensors, and estimating the object's position (and sometimes orientation) relative to the robot's reference frame; solving this perception task enables the robot to enact high-level actions like heading towards the object, avoiding it, or grasping it.  
Our work considers a specific visual object pose estimation case, in which the observing robot and a single target object are a Crazyflie nano-drone.

Object pose estimation subsystems typically rely on data from onboard stereo cameras~\cite{milella_model-based_nodate,schuster_distributed_2019}, lidars~\cite{schuppstuhl_annals_2022,wasik_lidar-based_2016}, infrared sensors~\cite{3d-indoor-flying}, ultra wideband (UWB) radios~\cite{shan_ultra-wideband_2022,9756977} or monocular vision~\cite{monocular-6d-ml,li_self-supervised_2022,frontnet}, being also our case.
Despite poor image camera quality, we demonstrate that it is sufficient to localize an observed \SI{10}{\centi\meter}-scale nano-drone, even without visual fiducial markers~\cite{monocular-6d-ml,s16050666,pickem_robotarium_2017}.

Depending on the application, one may need to estimate the full 3D pose of the object, i.e., its position and 3D orientation~\cite{schuster_distributed_2019,monocular-6d-ml}.
In this case, ad-hoc parameterizations can explicitly account for the topology of the SE(3) group of rigid transformations in a machine learning context~\cite{mahendran20173d}. 
In other applications, estimating only the 3D position is sufficient~\cite{3d-indoor-flying,nogps-3d-stereolaser-odom,infrared-indoor-3d,carrio_drone_2018,9756977}. 
In this paper, following previous work~\cite{frontnet}, we address the problem of drone pose estimation using its 3D position.

Visual pose estimation is a challenging pattern recognition problem and poses a significant computational burden, especially given the limited capabilities of our target nano-drone platform.
In some cases~\cite{offboard-ultravioletled}, such computation is offloaded to a more powerful, remote computer that receives sensory data and returns estimated poses.
Larger drones~\cite{depthmap-detection-jetson,flocking-outdoor-jetson} are equipped with powerful CPUs (sometimes including GPUs), and high-quality sensors. 
In contrast, one key aspect of our work is the integration of all computation aboard a memory/computation-limited Crazyflie nano-drone~\cite{ai-deck}, which poses significant challenges due to its microcontroller-class processors (sub-\SI{100}{\milli\watt}).

Nano-drones' localization systems often rely on UWB technology, employing either additional ad-hoc infrastructure~\cite{zhao2020learning} (UWB fixed anchors) or onboard multilateration algorithms~\cite{9756977}.
The former approach has the major disadvantages of minimal flexibility, deployability, and additional cost; the latter introduces a significant onboard power consumption overhead due to the UWB radio communication, up to $\sim$\SI{300}{\milli\watt} in active states~\cite{9756977}.
Differently, Li et al.~\cite{li_self-supervised_2022} recently tackled a vision-based pose estimation problem very similar to the one presented in this work and with the same hardware (i.e., drone, camera, MCUs).
Their contribution focuses on a novel approach to generating UWB-based training labels for a self-supervised CNN.
The UWB-based labels are used to train a CNN, which is then validated only offline on a limited 48-image testing set.
In contrast, our contribution lies in the system design, integration, and thorough experimental validation: our chosen architecture is larger (8 vs. 5 convolutional layers, 12$\times$ more parameters) yet can run 30\% faster as it requires fewer operations.
We also study real-time performance, provide an offline quantitative evaluation on a 15$\times$ larger testing set, and report a quantitative analysis of the in-field performance of the entire system integrated with a controller during uninterrupted flights.
Ultimately, our main contribution is application-agnostic, whereas our experimental validation explores a simple formation-flying task.

%% file: 03-implementation.tex
\section{System implementation} \label{sec:implementation}

\subsection{Robotic platform} \label{sec:prototype}

In this work, we employ the Bitcraze Crazyflie 2.1 open-source nano-drone, which features as main \textit{flight controller} an STM32 single-core microcontroller unit (MCU). 
In our configuration, we extend it with a commercial-off-the-shelf (COTS) Flow-deck to improve the onboard state estimation thanks to an optical-flow camera and time-of-flight sensor measuring the distance to the ground.
Additionally, we empower our nano-drone with an additional parallel ultra-low power (PULP) MCU, a QVGA monochrome camera, a WiFi radio, and 8/16\SI{}{\mega\byte} of DRAM/Flash memory, all hosted on the AI-deck COTS companion board.
Including all these additions, our nano-drones weights \SI{33}{\gram} and has a diameter of \SI{10}{\centi\meter}.

The PULP MCU is the Greenwaves Technologies GAP8\footnote{https://greenwaves-technologies.com/gap8\_mcu\_ai}, a RISC-V-based 9 cores processor.
The GAP8 has two power domains: the \textit{fabric controller} (FC), which features one core interacting with external memories/interfaces (e.g., DRAM/Flash and UART communication channel on the AI-deck), and with the second domain, the parallel \textit{cluster} (CL). 
The CL takes care of intensive computation using its 8-cores.
The overall memory hierarchy of this processor is organized into two layers; \SI{64}{\kilo\byte} of low-latency L1 memory shared among all cluster cores and \SI{512}{\kilo\byte} of L2 memory within the FC domain.
However, the GAP8 does not provide data caches or hardware floating-point units (FPUs), dictating explicit data management and the adoption of integer-quantized arithmetic, respectively.

\subsection{Training and testing datasets} \label{sec:Datasets}

Our work relies on labeled datasets acquired ad-hoc; two disjoint datasets are acquired for training and quantitative testing.
Each sample in a dataset is composed of: an input image ($320\times320$ gray-scale) and ground truth data for the four output variables, namely $x$, $y$, $z$, and $\phi$, i.e., the relative orientation of the target drone w.r.t. the observer one.
Ground truths are generated from the absolute poses of the flying drones, recorded with an 18-camera Optitrack infrared motion tracking setup.
During data acquisition, the drones are controlled by a ROS node via a predefined sequence of waypoints.
ROS also handles the precise matching between drones' poses and onboard images streamed via WiFi from the nano-drones.

To collect the training datasets, we leverage this setup to record multiple flights featuring different paths of the two drones, aiming at maximising pose variability for each component of the relative pose, and at recording images with a variety of backgrounds.
In particular, we use two motion patterns.
In one pattern, the two drones rotate around the center of the room while facing each other; both drones record data simultaneously, thus acting as both an observer and a target.  
Multiple flights are repeated while modulating: the radius of the circle that the two drones follow; the relative height between the two drones; and the angular offset of the two drones along the circle.
In another pattern, the observer drone is static while the target flies along circles lying on a vertical plane, orthogonal to the optical axis of the observer; the target follows multiple of these circles, at different distances from the observer and with different radii.

After data collection, all samples for which the target drone is not in the field of view of the observer are discarded; we obtain a total of 21001 samples, which we randomly divided in 19664 and 1337 samples respectively for training and validation.
The images are augmented on-the-fly during the training phase.
Random augmentations affect image exposure, gamma, and dynamic range; in addition, noise, blur and/or vignetting are randomly applied to each image.

\begin{figure}[tbh]
  \includegraphics[width=\columnwidth]{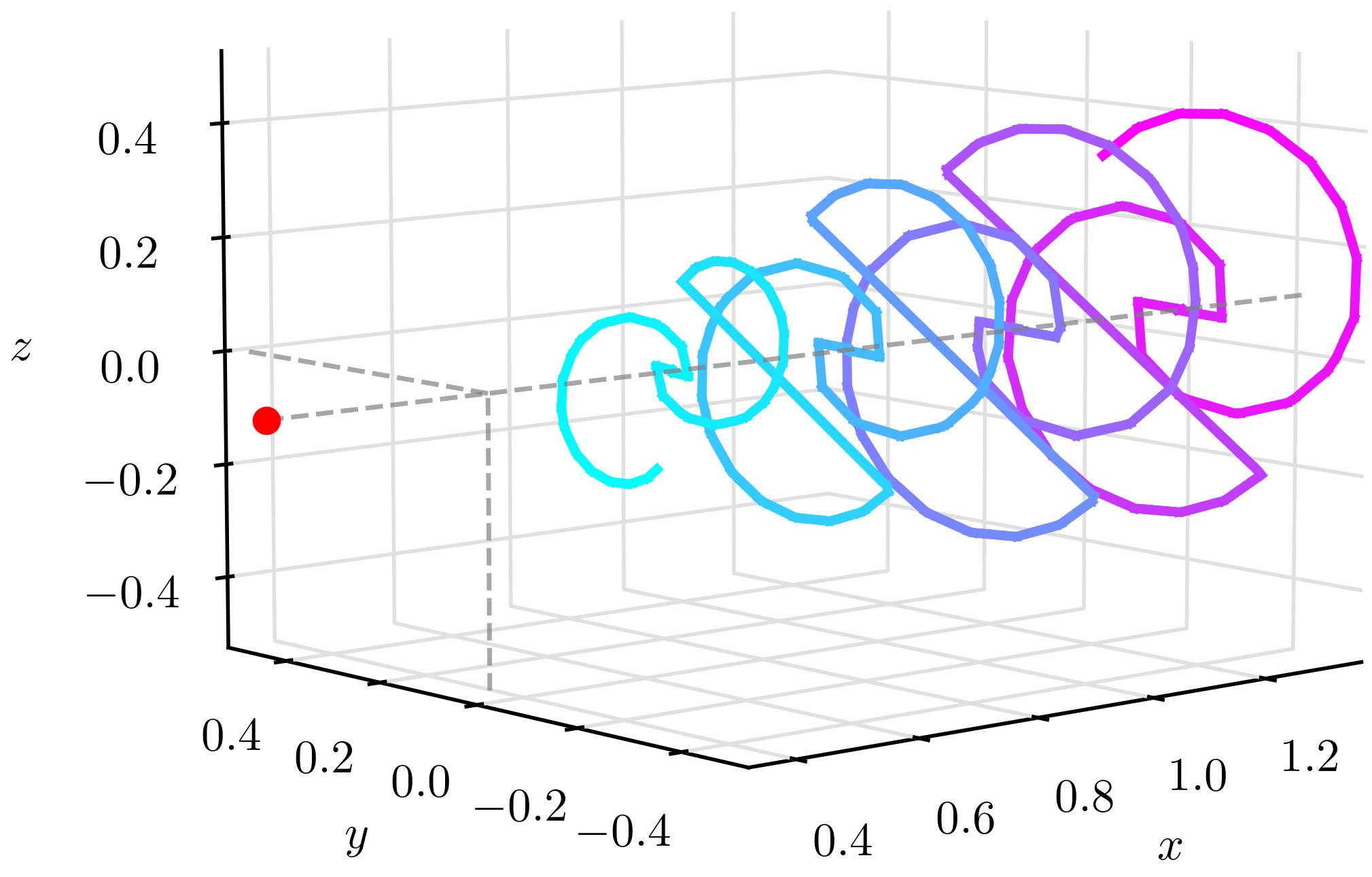}
  \caption{3D trajectory flown by the target nano-drone to collect our testing dataset. The observer drone, shown in red, is placed at $(x,y,z)=(0,0,0)$ pointing towards positive $x$}
  \label{fig:spiral_trajectory}
\end{figure}

For the testing dataset, we recorded a new session with a behavior that is not represented in any of the training datasets; this ensures that our quantitative evaluation does not reward any potential overfitting on the training data.
In particular, the target drone follows a spiral-like trajectory, which enforces significant variability for all components of the relative pose.
The drones are initially aligned on the $x$ axis, \SI{0.2}{\meter} apart and facing each other.
While the observer drone is static, the target drone slowly flies farther away from the observer along the $x$ axis.
While doing so, it alternates inward and outward spiral motions on said axis, as depicted in Figure~\ref{fig:spiral_trajectory}.
This procedure generated a total of 754 samples.

\subsection{Deep neural network} \label{sec:DNN}

Two CNNs are tested: PULP-Frontnet, a small-sized network proofed for perception tasks and nano-drones, and MobileNetV2, a popular network for edge devices. 
In the case of Frontnet, the network starts with a $5\times5$ convolutional layer, followed by a $2\times2$ max-pooling layer, and then three repetitions of a custom block made of two $3\times3$ convolutions.
The first convolution layer of each block has a stride of 2, further reducing the feature maps size and increasing the number of output channels.
All convolutional layers are followed by a batch-normalization layer and a ReLU activation.

\begin{figure*}[tb]
  \includegraphics[width=\textwidth]{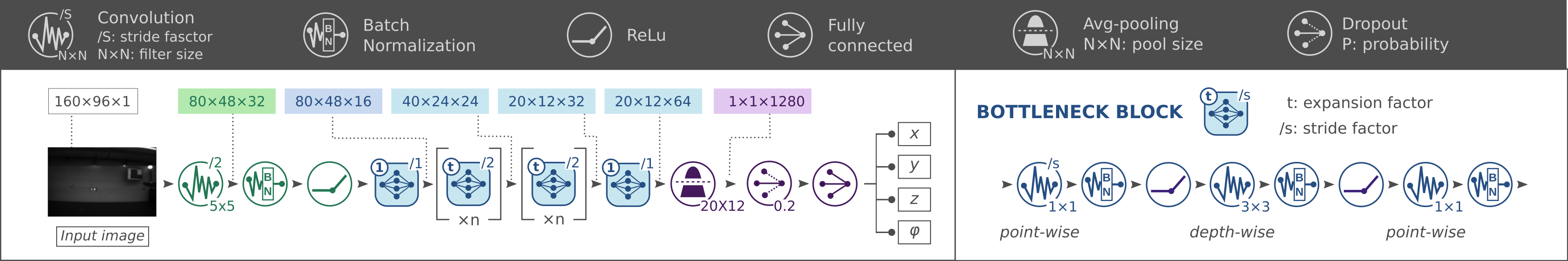}
  \caption{Template of the network structure used for our lightweight MobileNetV2 variants.}
  \label{fig:mobilenet}
\end{figure*}

In the case of MobileNetV2, we consider 16 variants that are lighter than the original architecture in terms of both memory and computational needs.
These networks consist of a standard convolutional layer, four inverted residual blocks, an average pooling layer and a fully connected layer.
Figure~\ref{fig:mobilenet} shows the template of our reduced MobileNetV2.
Our search space is parametrized by $t$ and $n$, which refer to the expansion factor and the number of repetitions of the two intermediate blocks, respectively.
We consider the 15 models defined as $(t,n) \in \{6,8,10,12,14\} \times \{2,3,4\}$, where $\times$ denotes the cartesian product; we further include the case $(t,n) = (2, 2)$, i.e. the smallest possible variant considering our stride layout constraints.

\subsection{System deployment} \label{sec:control}

Once the models are trained in PyTorch, we need to change their numerical representation from a 32-bit floating point representation (i.e., full precision) to an 8-bit integer fixed-point arithmetics due to the missing FPU in our target processor (Section~\ref{sec:prototype}).
We use the NEMO~\cite{NEMO} library, which uses parametrized clipping activation technique~\cite{choi2018pact}, to quantize the CNN's weights and activations in three sequential steps.
The first one trains a floating-point CNN to minimize the sum of the L1 loss for the pose estimation vector ($x,y,z,\phi$).
Then, we convert the model with the lower validation loss to the so-called  \textit{fake-quantized} version, where weights and activations magnitude can assume only a discrete (0-255) set of values, still using float32 data-types.
Before moving to the last stage, we perform a fine-tuning PACT-based quantization stage over ten epochs.
We conclude the process with the \textit{integer deployable} stage, where all tensors are represented by integer numbers and a floating point scale factor.
At this point, the network can be executed entirely by multiplying and accumulating only integer values.

The second tool employed is called DORY~\cite{dory}, which, starting from the quantized model, produces optimized C code.
By combining optimized basic kernels (e.g., convolutions, linear layers, max-pooling layers, etc.) in a template-based wrapping code, DORY also takes care of the memory orchestration to exploit data-locality in the GAP8's L2 and L1 memories and off-chip DRAM.

To complete the development of our pipeline, we extend the inference routine generated by DORY, with a \textit{i}) image acquisition routine, \textit{ii}) image cropping (from $160\times160$ to $160\times96$pixels), and \textit{iii}) the final UART communication to forward the CNN's output to the main STM32 MCU aboard the nano-drone.
Finally, our pipeline overlaps the inference time (the longest of all routines) with image acquisition and cropping (together $\sim$ \SI{0.6}{\mega cycles}) in a double-buffered fashion.
With this mechanism, the FC can start acquiring and cropping a new image when the CL is computing the previous one, limiting the overall pipeline's latency only to the inference routine.

Once a new prediction ($x,y,z,$ and $\phi$) is sent from the GAP8 to the STM32, the pose is filtered by a Kalman filter, smoothing the sequence of poses.
The output of the filter is fed to a velocity controller, as in~\cite{frontnet}, that converts this temporally-consistent series of poses into velocity setpoints to keep the target nano-drone at a constant distance of \SI{0.8}{\meter}, and then forwarded to the Crazyflie PID-cascade controller.

%% file: 04-results.tex
\section{Results} \label{sec:results}

\subsection{Comparison of network architectures} \label{sec:dnn_comparison}

\begin{figure*}[tb]
  \includegraphics[width=\textwidth]{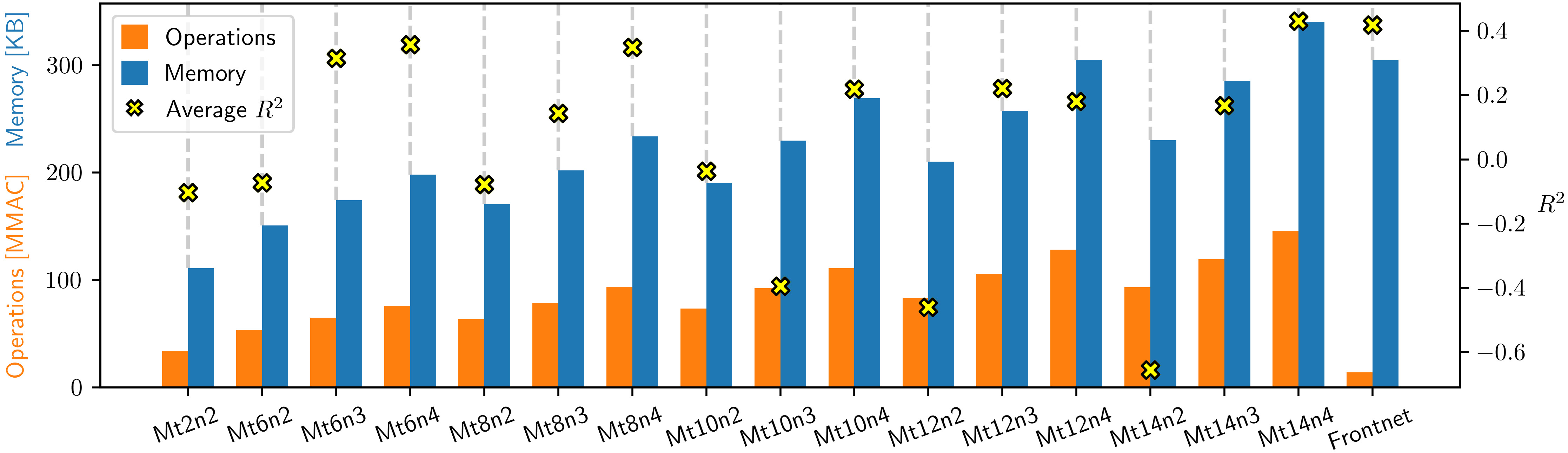}
  \caption{Overview of all tested architectures; bars refer to the left axis and denote the number of MAC operations (orange) and number of 8-bit weights (blue);  crosses refer to the right axis and denote the $R^2$ score averaged over the $x$, $y$ and $z$ outputs.}
  \label{fig:CNNs_overview}
\end{figure*}

In this section, we compare the PULP-Frontnet model with 16 variants of the MobileNetV2 CNN.
Figure~\ref{fig:CNNs_overview} reports three metrics for each architecture: the memory footprint (i.e., number of 8-bit weights); the number of multiply-accumulate (MAC) operations, which directly affects the achievable throughput in terms of frames per second; and the regression performance.

We quantify regression performance separately for each output variable using the $R^2$ score (i.e., the \emph{coefficient of determination}), a standard metric~\cite{R2_note_91} that measures the proportion of the variation in the output that the model explains.
A dummy predictor that always outputs the average of the testing samples scores $R^2=0$; instead, an ideal regressor yields $R^2=1$.
Figure~\ref{fig:CNNs_overview} reports the $R^2$ score averaged over the $x,y,$ and $z$ components; it ignores the performance on the $\phi$ component, which is poor for all models, as we discuss below. 

The best regression performance is given by the largest MobileNetV2 model, which is closely followed by PULP-Frontnet ($R^2 = 0.43$ vs. $0.42$, respectively).
We observe that MobileNetV2 architectures show a consistent pattern and yield better regression performance for larger values of $n$.
This pattern suggests that deeper MobileNetV2 models are more favorable for our problem than wider ones with the same number of parameters.

In terms of memory required to store weights, MobileNetV2 models span from \SI{111}{\kilo\byte} to \SI{340}{\kilo\byte}, while the PULP-Frontnet model requires \SI{304}{\kilo\byte}; all such models fit the L2 memory of the GAP8 SoC.
The difference in terms of MACs is much more relevant, with all MobileNetV2 models requiring significantly more operations than PULP-Frontnet, up to 10.4$\times$.  
Assuming a computational efficiency of 4 MAC/cycle for all models and a CL frequency of \SI{170}{\mega\hertz}, this would result in \SI{20}{frame/\second} and \SI{4.6}{frame/\second}, for the smallest and the largest MobileNetV2 respectively, while PULP-Frontnet would score \SI{48}{frame/\second}.
Therefore, we select PULP-Frontnet as our candidate to be deployed and field-tested on the nano-drone.

\begin{figure}[tb]
  \includegraphics[width=\columnwidth]{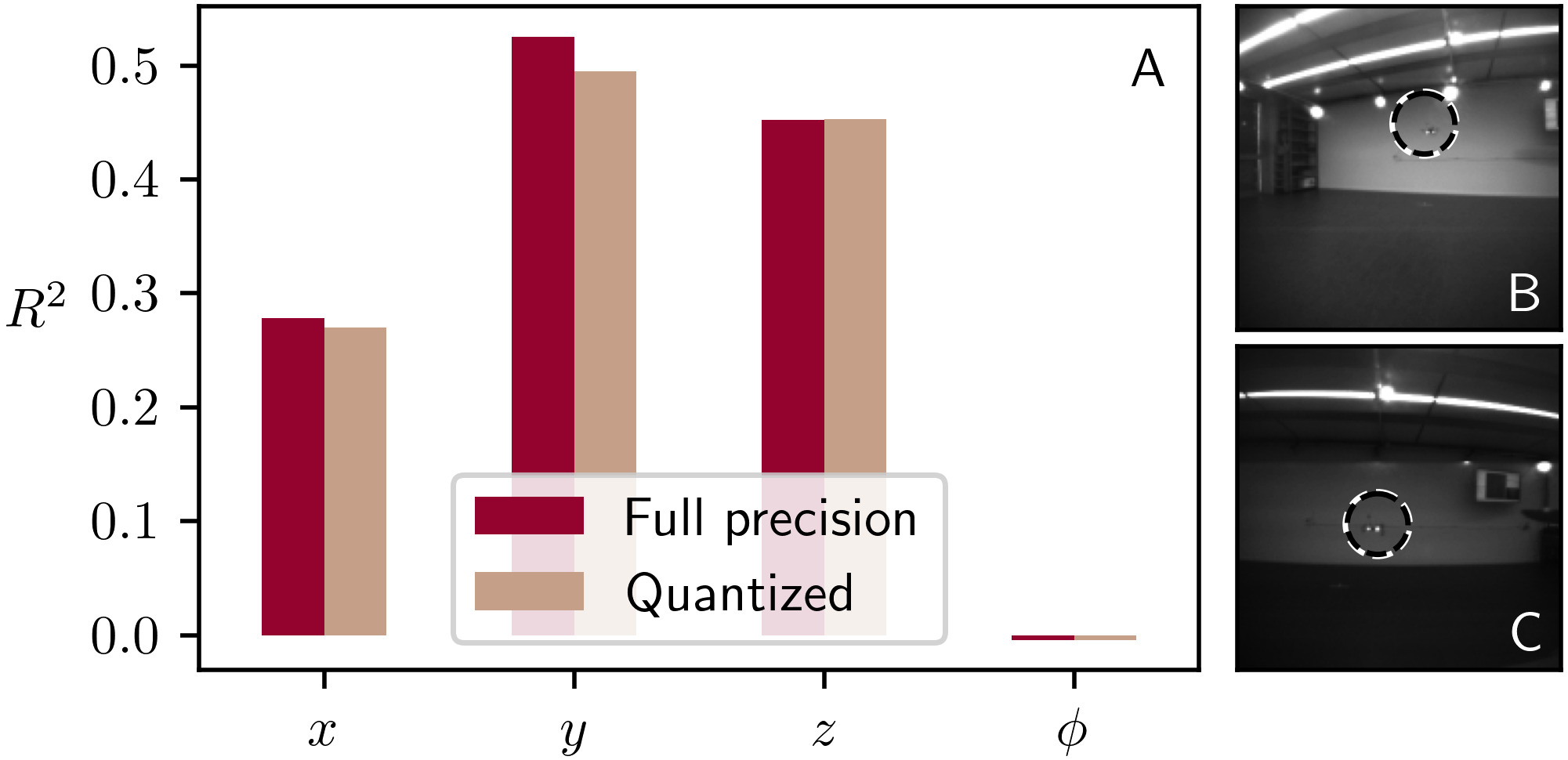}
  \caption{A) Regression performance of PULP-Frontnet for each output variable; full-precision (red) and quantized (tan) models.  B,C) two images with different $\phi$ (target drone is circled).}
  \textbf{\label{fig:r2_frontnet}}
\end{figure}

In Figure~\ref{fig:r2_frontnet}-A, we report the $R^2$ score for each output variable of the PULP-Frontnet CNN, including both full precision and quantized versions.
$x$, $y$, and $z$ outputs yield a good regression performance up to $R^2=0.53$, while the $\phi$ component scores nearly 0, i.e., does not outperform a trivial predictor.
Such poor performance on the $\phi$ component is expected: correctly predicting $\phi$ implies understanding the nano-drone orientation through low-resolution, monochrome and noisy images; it is a next-to-impossible task also for human observers, since the target appearance is not significantly affected by its orientation, as shown in Figure~\ref{fig:r2_frontnet}-B/C.
Figure~\ref{fig:r2_frontnet}-A also shows that the post-training quantization stage has a negligible impact on regression performance: $R^2$ drops by at most 3 percentage points (0.52 to 0.49).


\subsection{Regression performance} \label{sec:reg_perf}

\begin{figure*}[tb]
  \includegraphics[width=\textwidth]{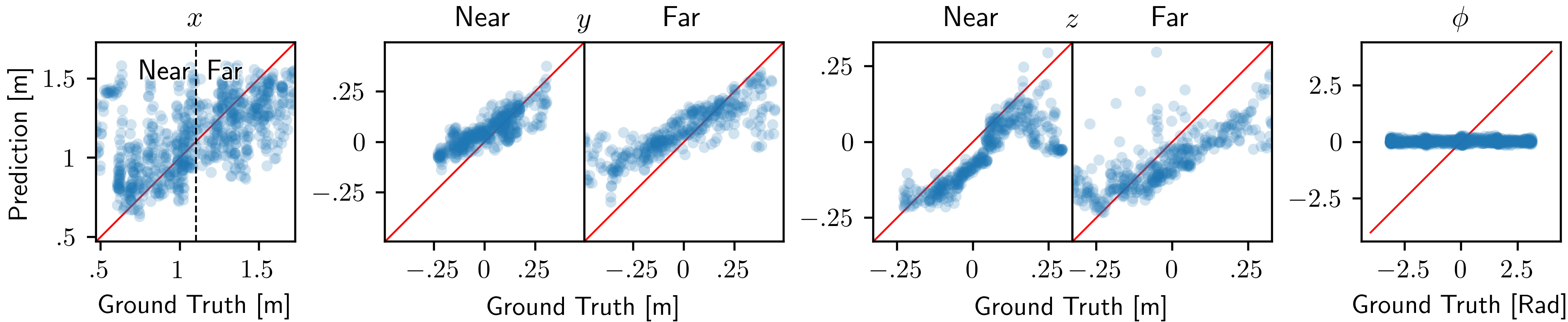}
  \caption{
  Predictions (ordinates axis) vs ground truth (abscissae axis) for each of the output variables, for the entire testing set (one point per sample). For $y$, $z$ and $\phi$ outputs we separately report samples in which the target is near ($x<$\SI{1.1}{\meter}) and far ($x\geq$\SI{1.1}{\meter}).
  }
  \label{fig:scatterplot_testing}
\end{figure*}

Figure~\ref{fig:scatterplot_testing} further analyzes how the PULP-Frontnet model predictions compare to ground truth for each output variable; an ideal estimator would yield a scatterplot in which all points lie on the diagonal line.

We first observe that the prediction of $x$ (i.e., the target's distance) is challenging.
This is expected as $x$ can only be inferred by the size of the target's image, which only covers a few pixels and is, therefore, difficult to assess for the model (and for a human observer).
Outputs $y$ and $z$ are estimated with significantly higher accuracy; these variables are related to the position of the target's image on the horizontal and vertical axis, respectively: easier to predict.
We further observe that when the target is far, the predictions tend to be more conservative, i.e., closer to 0.  
This occurs because the model is not confident about the target's position since its image becomes smaller and harder to detect.  Furthermore, the $y$ and $z$ components of the relative pose depend on the position of the target's image in the frame but also on the target's distance; therefore, the uncertainty that affects the distance estimation is mirrored in the predictions for $y$ and $z$. 

On the $z$ component, i.e., the relative height between the two drones, we observe that the predictions are slightly, but consistently, underestimated (by $\sim$\SI{5}{\centi\meter}).
We correlate this effect to the differences in the training and testing datasets' data distribution or a slight misalignment in the camera pitch on the quadrotor.
The rightmost plots of Figure~\ref{fig:scatterplot_testing} show that the model, as previously explained, does not return any useful information for $\phi$, i.e., the orientation component of the relative pose.

\subsection{Onboard real-time performance} \label{sec:onboard_perf}

This section presents the onboard performance evaluation of the PULP-Frontnet CNN running at different operative points, i.e., voltage and frequencies.
As introduced in~\cite{frontnet}, we explore three configurations, namely \textit{minimum power} (VDD@\SI{1.0}{\volt} FC@\SI{25}{\mega\hertz} CL@\SI{25}{\mega\hertz}), \textit{most energy efficient} (VDD@\SI{1.0}{\volt} FC@\SI{25}{\mega\hertz} CL@\SI{75}{\mega\hertz}), and \textit{maximum performance} (VDD@\SI{1.2}{\volt} FC@\SI{250}{\mega\hertz} CL@\SI{175}{\mega\hertz}).
The minimum power is particularly relevant for all those scenarios aiming at maximizing the system's lifetime, for example, when the nano-drone lands and starts behaving as an ubiquitous smart-sensor.
Instead, the most energy-efficient configuration should always be preferred if its throughput fulfills the mission's requirements.
Lastly, the maximum performance operative point should be considered to maximize the CNN's inference rate, for example, when agility and responsiveness are keys.

\begin{table}[tb]
\centering
\caption{PULP-Frontnet throughput and power consumption.}
\label{tab:power_cycles}
\begin{tabular}{@{}llll@{}}
\toprule
Frequencies [\SI{}{\mega\hertz}] & VDD [\SI{}{\volt}] & Frame-rate [\SI{}{fps}] & Power [\SI{}{\milli\watt}] \\ \midrule
FC/CL @ 25/25                    & 1.0                & 6.8                     & 9.9                        \\
FC/CL @ 25/75                    & 1.0                & 19.7                    & 25.1                       \\
FC/CL @ 250/175                  & 1.2                & 48.3                    & 95.4                       \\ \bottomrule
\end{tabular}
\end{table}

\begin{figure}[tb]
  \includegraphics[width=\columnwidth]{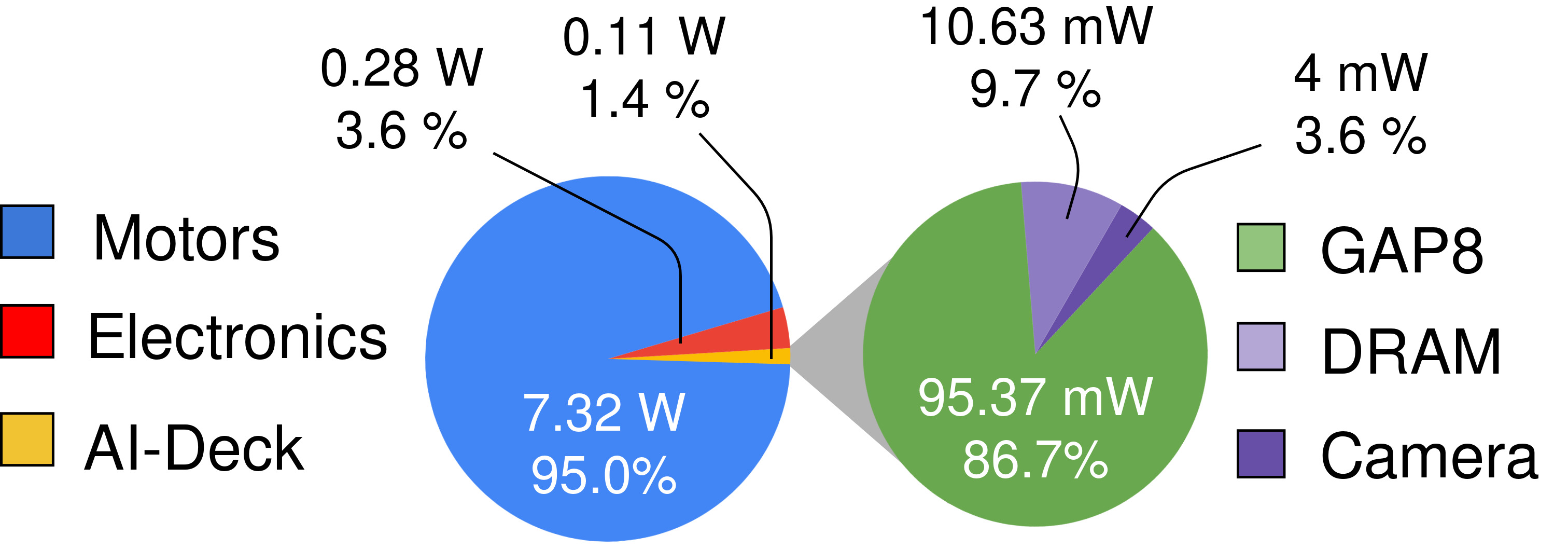}
  \caption{Robotic platform's power breakdown running PULP-Frontnet at VDD@\SI{1.2}{\volt}, FC@\SI{250}{\mega\hertz}, and CL@\SI{175}{\mega\hertz}.}
  \label{fig:power_breakdown}
\end{figure}

Table~\ref{tab:power_cycles} reports performance, in frames-per-second (fps), and mean power consumption profiling the GAP8 SoC with a RocketLogger data logger~\cite{sigrist2016rocketlogger} (\SI{64}{\kilo sp\second}).
Results span from a minimum of \SI{6.8}{fps}@\SI{9.9}{\milli\watt} up to \SI{48.3}{fps}@\SI{95.4}{\milli\watt}.
The overall nano-drone power breakdown, in Figure~\ref{fig:power_breakdown}, shows the expected trends, with the actuators' power consumption dominating (95\% of the total) and leaving a small fraction (5\%) to the electronics.
Finally, the GAP8 SoC, running our monocular relative localization CNN accounts only for the 1.2\% of the total, well aligned with the SoA~\cite{frontnet,9606685}.

\subsection{In-field control performance} \label{sec:control_perf}

\begin{figure}[tb]
  \includegraphics[width=\columnwidth]{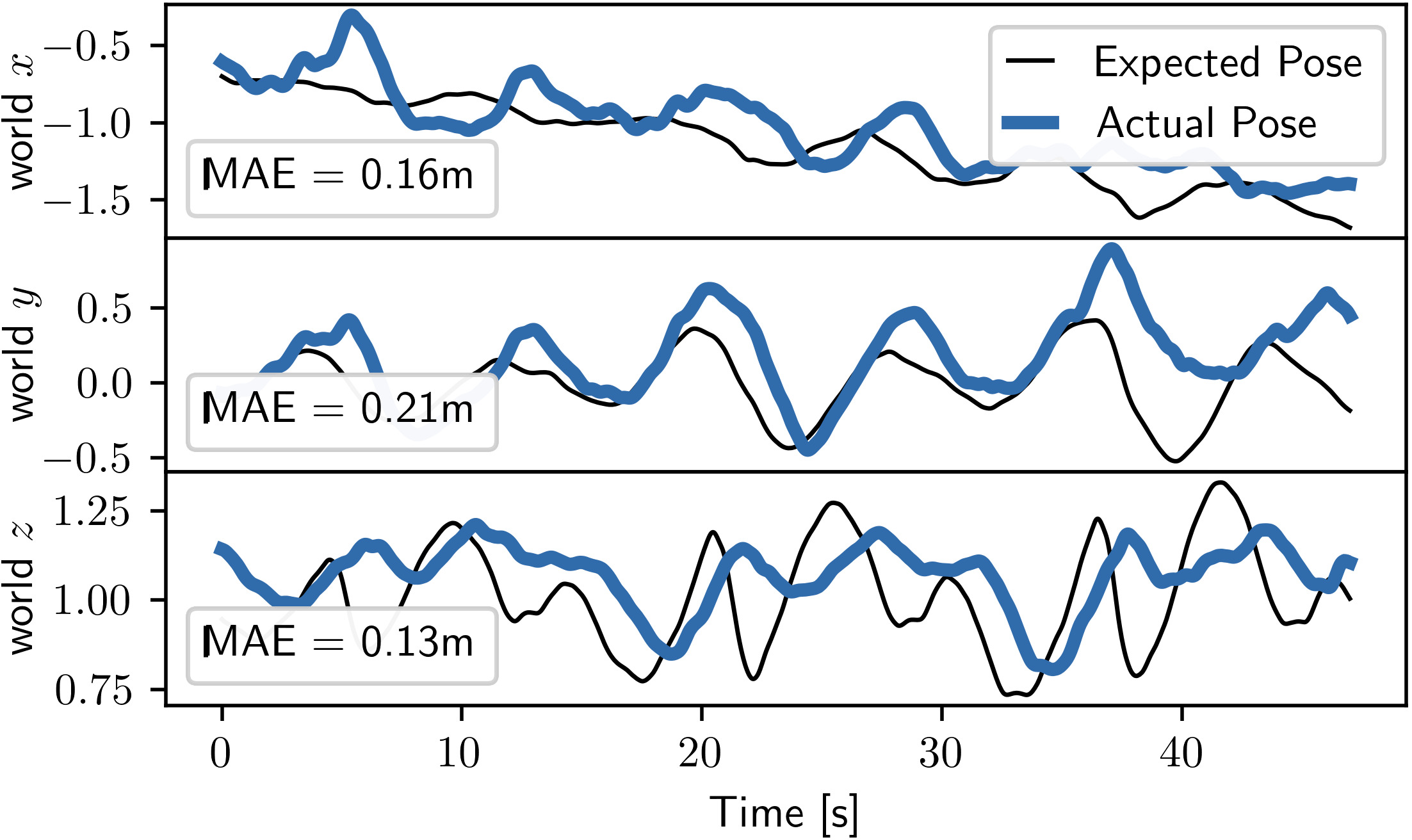}
  \caption{The $x$, $y$ and $z$ components of the absolute world pose of the observer drone during in-field tests. The thick blue line represents the true pose of the drone; the thin black line is the ideal pose that the drone is expected to track, i.e. a fixed relative pose from the moving target.}
  \label{fig:spiral_control}
\end{figure}

We perform an in-field test with two flying drones to prove our system.
One drone takes the role of the target and performs pre-defined movements, following the same spiral-like trajectory described in Section~\ref{sec:Datasets}. 
The other drone (observer) tries to keep a fixed relative pose w.r.t. to the target drone: a distance of \SI{1}{\meter} ($x = 0.8$m); the same height ($z = 0$m); and a position that keeps the target in the center of the frame ($y = 0$m) while keeping the observer's yaw fixed. 
Our CNN runs in real-time on the observer nano-drone, and its predictions are used for closed-loop control.
During this experiment, the pose of both drones is recorded by a motion capture system.  
The observer drone is oriented in such a way that its $x$ axis (i.e., the forward direction) is aligned with the world's $x$ axis.  
For quantitative performance evaluation, we compare the observed drone's recorded pose (in world coordinates) with the expected one, as shown in Figure~\ref{fig:spiral_control}.
We also include the mean absolute error (MAE) scores for an easier comparison of the three components.

The observer can reliably estimate the relative position for the entire test ($\sim$\SI{60}{\second}), as shown in the supplementary video.
Despite the small latency due to the controller computation, we can observe a close match in absolute coordinates for all the components. 
The $x$ and $z$ components are the most accurate, with a MAE of \SI{16}{\centi\meter} and \SI{13}{\centi\meter}, respectively.
Instead, on the $y$ output, we see that the observer slightly overshoots whenever it needs to move to the left, resulting in a MAE of \SI{21}{\centi\meter}
On the other hand, we can observe that the responsiveness on this component is the highest, with the drone swiftly inverting the $y$ direction when needed.
Targeting a nano-drone pose estimation task is highly challenging as we need to estimate the pose of a tiny object, which is as big as $\sim8\times18$ pixels at a distance of \SI{0.4}{\meter} or $\sim2\times5$ pixels at \SI{1.5}{\meter}, i.e., 0.9\% and 0.06\% of the input image, respectively.
Comparing these results with the SoA PULP-Frontnet baseline~\cite{frontnet}, targeting human pose estimation, the same level of complexity would require a human subject more than 60 meters away from the nano-drone.


We also tested the model in a scenario where the target drone performs a random path instead of the predefined one. 
The observer successfully locks and holds its relative pose w.r.t. the target drone for the entire duration of a 2-minutes experiment.
This experiment is representative of swarm operations where peer drones need to establish line-of-sight conditions.
The available supplementary video material shows the ability of the drone to continuously maintain line of sight for approximately half of the drone's lifetime.

Finally, to stress the generalization capabilities of our final CNN, we perform one additional in-field test in a \textit{never-seen-before} indoor environment, which is different from both previous sections' experiments and not present in any dataset.  
The target drone is remotely controlled by a human pilot along 3D trajectories spanning a volume of approximately $4\times4\times$\SI{3}{\meter}.
The autonomous observer drone starts its mission at a distance of \SI{1.2}{\meter} from the target drone.
The experiment results are reported in the supplementary video.
Since this environment (i.e., student room) is not equipped with a motion tracking system, we can not compute quantitative metrics on the position tracking performance.  
The observer drone behaves as expected and successfully tracks the target and keeps it in the field of view for a duration of \SI{53}{\second} while maintaining a set relative distance of $x$=\SI{0.8}{\meter}.
In contrast, depending on the run, a non-moving observer would have lost view of the target after a timespan of $\sim$2 to \SI{10}{\second}.

%% file: 05-conclusion.tex
\section{Conclusion} \label{sec:conclusion}

This paper addressed the perceptive problem of monocular relative pose estimation between two resource-limited nano-drones.
We presented a vertically-integrated system based on deep learning using only visual information.
After characterizing multiple (i.e., 17) CNNs, we select the PULP-Frontnet model to run aboard our nano-drone.
We cope with the end-to-end development and deployment pipeline, from the dataset collection, augmentation, quantization, and system optimizations.
In summary, our results demonstrate precise localization (average control error of \SI{17}{\centi\meter}) of a \SI{10}{\centi\meter}-size target nano-drone by employing only low-resolution monochrome images, up to $\sim$\SI{2}{\meter} distance, with long-endurance in-field performance, as much as \SI{2}{\minute} experiments.

%% file: 00-main.bbl
\begin{thebibliography}{10}
\providecommand{\url}[1]{#1}
\csname url@rmstyle\endcsname
\providecommand{\newblock}{\relax}
\providecommand{\bibinfo}[2]{#2}
\providecommand\BIBentrySTDinterwordspacing{\spaceskip=0pt\relax}
\providecommand\BIBentryALTinterwordstretchfactor{4}
\providecommand\BIBentryALTinterwordspacing{\spaceskip=\fontdimen2\font plus
\BIBentryALTinterwordstretchfactor\fontdimen3\font minus
  \fontdimen4\font\relax}
\providecommand\BIBforeignlanguage[2]{{%
\expandafter\ifx\csname l@#1\endcsname\relax
\typeout{** WARNING: IEEEtran.bst: No hyphenation pattern has been}%
\typeout{** loaded for the language `#1'. Using the pattern for}%
\typeout{** the default language instead.}%
\else
\language=\csname l@#1\endcsname
\fi
#2}}

\bibitem{li_self-supervised_2022}
S.~Li, C.~De~Wagter, and G.~C. H.~E. De~Croon, ``Self-supervised {Monocular}
  {Multi}-robot {Relative} {Localization} with {Efficient} {Deep} {Neural}
  {Networks},'' in \emph{2022 {International} {Conference} on {Robotics} and
  {Automation} ({ICRA})}, May 2022, pp. 9689--9695.

\bibitem{palossi2017target}
D.~Palossi, J.~Singh, M.~Magno, and L.~Benini, ``Target following on nano-scale
  unmanned aerial vehicles,'' in \emph{2017 7th IEEE international workshop on
  advances in sensors and interfaces (IWASI)}.\hskip 1em plus 0.5em minus
  0.4em\relax IEEE, 2017, pp. 170--175.

\bibitem{frontnet}
D.~Palossi, N.~Zimmerman, A.~Burrello, F.~Conti, H.~Müller, L.~M. Gambardella,
  L.~Benini, A.~Giusti, and J.~Guzzi, ``Fully onboard {AI}-powered human-drone
  pose estimation on ultra-low power autonomous flying nano-{UAVs},''
  \emph{IEEE Internet of Things Journal}, pp. 1--1, 2021.

\bibitem{monocular-6d-ml}
A.~Breitenmoser, L.~Kneip, and R.~Siegwart, ``A monocular vision-based system
  for 6d relative robot localization,'' in \emph{2011 IEEE/RSJ International
  Conference on Intelligent Robots and Systems}, 2011, pp. 79--85.

\bibitem{carrio_drone_2018}
A.~Carrio, S.~Vemprala, A.~Ripoll, S.~Saripalli, and P.~Campoy, ``Drone
  detection using depth maps,'' in \emph{2018 IEEE/RSJ International Conference
  on Intelligent Robots and Systems (IROS)}, 2018, pp. 1034--1037.

\bibitem{schilling_vision-based_2021}
F.~Schilling, F.~Schiano, and D.~Floreano, ``Vision-based drone flocking in
  outdoor environments,'' \emph{IEEE Robotics and Automation Letters}, vol.~6,
  no.~2, pp. 2954--2961, 2021.

\bibitem{milella_model-based_nodate}
A.~Milella, F.~Pont, and R.~Siegwart, ``\BIBforeignlanguage{en}{Model-{Based}
  {Relative} {Localization} for {Cooperative} {Robots} {Using} {Stereo}
  {Vision}},'' \emph{\BIBforeignlanguage{en}{-}}, p.~8, 2005.

\bibitem{schuster_distributed_2019}
\BIBentryALTinterwordspacing
M.~J. Schuster, K.~Schmid, C.~Brand, and M.~Beetz,
  ``\BIBforeignlanguage{en}{Distributed stereo vision-based {6D} localization
  and mapping for multi-robot teams},'' \emph{\BIBforeignlanguage{en}{J Field
  Robotics}}, vol.~36, no.~2, pp. 305--332, Mar. 2019. [Online]. Available:
  \url{https://onlinelibrary.wiley.com/doi/10.1002/rob.21812}
\BIBentrySTDinterwordspacing

\bibitem{schuppstuhl_annals_2022}
\BIBentryALTinterwordspacing
T.~Schüppstuhl, K.~Tracht, and A.~Raatz, Eds.,
  \emph{\BIBforeignlanguage{en}{Annals of {Scientific} {Society} for
  {Assembly}, {Handling} and {Industrial} {Robotics} 2021}}.\hskip 1em plus
  0.5em minus 0.4em\relax Cham: Springer International Publishing, 2022.
  [Online]. Available:
  \url{https://link.springer.com/10.1007/978-3-030-74032-0}
\BIBentrySTDinterwordspacing

\bibitem{wasik_lidar-based_2016}
A.~Wasik, R.~Ventura, J.~N. Pereira, P.~U. Lima, and A.~Martinoli,
  ``Lidar-{Based} {Relative} {Position} {Estimation} and {Tracking} for
  {Multi}-robot {Systems},'' in \emph{Robot 2015: {Second} {Iberian} {Robotics}
  {Conference}}, L.~P. Reis, A.~P. Moreira, P.~U. Lima, L.~Montano, and
  V.~Muñoz-Martinez, Eds.\hskip 1em plus 0.5em minus 0.4em\relax Cham:
  Springer International Publishing, 2016, pp. 3--16.

\bibitem{3d-indoor-flying}
\BIBentryALTinterwordspacing
J.~F. Roberts, T.~Stirling, J.-C. Zufferey, and D.~Floreano, ``3-d relative
  positioning sensor for indoor flying robots,'' \emph{Autonomous Robots},
  vol.~33, no.~1, pp. 5--20, Aug 2012. [Online]. Available:
  \url{https://doi.org/10.1007/s10514-012-9277-0}
\BIBentrySTDinterwordspacing

\bibitem{shan_ultra-wideband_2022}
F.~Shan, H.~Huo, J.~Zeng, Z.~Li, W.~Wu, and J.~Luo, ``Ultra-{Wideband} {Swarm}
  {Ranging} {Protocol} for {Dynamic} and {Dense} {Networks},'' \emph{IEEE/ACM
  Transactions on Networking}, pp. 1--15, 2022, conference Name: IEEE/ACM
  Transactions on Networking.

\bibitem{9756977}
V.~Niculescu, D.~Palossi, M.~Magno, and L.~Benini, ``Energy-efficient, precise
  uwb-based 3-d localization of sensor nodes with a nano-uav,'' \emph{IEEE
  Internet of Things Journal}, pp. 1--1, 2022.

\bibitem{s16050666}
\BIBentryALTinterwordspacing
F.~Vanegas and F.~Gonzalez, ``Enabling uav navigation with sensor and
  environmental uncertainty in cluttered and gps-denied environments,''
  \emph{Sensors}, vol.~16, no.~5, 2016. [Online]. Available:
  \url{https://www.mdpi.com/1424-8220/16/5/666}
\BIBentrySTDinterwordspacing

\bibitem{pickem_robotarium_2017}
D.~Pickem, P.~Glotfelter, L.~Wang, M.~Mote, A.~Ames, E.~Feron, and
  M.~Egerstedt, ``The {Robotarium}: {A} remotely accessible swarm robotics
  research testbed,'' in \emph{2017 {IEEE} {International} {Conference} on
  {Robotics} and {Automation} ({ICRA})}, May 2017, pp. 1699--1706.

\bibitem{mahendran20173d}
S.~Mahendran, H.~Ali, and R.~Vidal, ``3d pose regression using convolutional
  neural networks,'' in \emph{Proceedings of the IEEE International Conference
  on Computer Vision Workshops}, 2017, pp. 2174--2182.

\bibitem{nogps-3d-stereolaser-odom}
\BIBentryALTinterwordspacing
M.~e.~a. Achtelik, ``Stereo vision and laser odometry for autonomous
  helicopters in gps-denied indoor environments.'' \emph{The International
  Society for Optical Engineering}, 2009. [Online]. Available:
  \url{https://dspace.mit.edu/handle/1721.1/52660}
\BIBentrySTDinterwordspacing

\bibitem{infrared-indoor-3d}
N.~Kirchner and T.~Furukawa, ``Infrared localisation for indoor uavs,''
  \emph{-}, 01 2005.

\bibitem{offboard-ultravioletled}
V.~Walter, M.~Saska, and A.~Franchi, ``Fast mutual relative localization of
  uavs using ultraviolet led markers,'' in \emph{2018 International Conference
  on Unmanned Aircraft Systems (ICUAS)}, 2018, pp. 1217--1226.

\bibitem{depthmap-detection-jetson}
A.~Carrio, S.~Vemprala, A.~Ripoll, S.~Saripalli, and P.~Campoy, ``Drone
  detection using depth maps,'' in \emph{2018 IEEE/RSJ International Conference
  on Intelligent Robots and Systems (IROS)}, 2018, pp. 1034--1037.

\bibitem{flocking-outdoor-jetson}
F.~Schilling, F.~Schiano, and D.~Floreano, ``Vision-based drone flocking in
  outdoor environments,'' \emph{IEEE Robotics and Automation Letters}, vol.~6,
  no.~2, pp. 2954--2961, 2021.

\bibitem{ai-deck}
D.~Palossi, F.~Conti, and L.~Benini, ``An open source and open hardware deep
  learning-powered visual navigation engine for autonomous nano-uavs,'' in
  \emph{2019 15th International Conference on Distributed Computing in Sensor
  Systems (DCOSS)}, 2019, pp. 604--611.

\bibitem{zhao2020learning}
W.~Zhao, A.~Goudar, J.~Panerati, and A.~P. Schoellig, ``Learning-based bias
  correction for ultra-wideband localization of resource-constrained mobile
  robots,'' \emph{arXiv preprint arXiv:2003.09371}, 2020.

\bibitem{NEMO}
\BIBentryALTinterwordspacing
F.~Conti, ``Technical report: {NEMO} {DNN} quantization for deployment model,''
  \emph{CoRR}, vol. abs/2004.05930, 2020. [Online]. Available:
  \url{https://arxiv.org/abs/2004.05930}
\BIBentrySTDinterwordspacing

\bibitem{choi2018pact}
J.~Choi, Z.~Wang, S.~Venkataramani, P.~I.-J. Chuang, V.~Srinivasan, and
  K.~Gopalakrishnan, ``Pact: Parameterized clipping activation for quantized
  neural networks,'' \emph{arXiv preprint arXiv:1805.06085}, 2018.

\bibitem{dory}
A.~Burrello, A.~Garofalo, N.~Bruschi, G.~Tagliavini, D.~Rossi, and F.~Conti,
  ``Dory: Automatic end-to-end deployment of real-world dnns on low-cost iot
  mcus,'' \emph{IEEE Transactions on Computers}, vol.~70, no.~8, pp.
  1253--1268, 2021.

\bibitem{R2_note_91}
N.~J. Nagelkerke \emph{et~al.}, ``A note on a general definition of the
  coefficient of determination,'' \emph{Biometrika}, vol.~78, no.~3, pp.
  691--692, 1991.

\bibitem{sigrist2016rocketlogger}
L.~Sigrist, A.~Gomez, R.~Lim, S.~Lippuner, M.~Leubin, and L.~Thiele,
  ``Rocketlogger: Mobile power logger for prototyping iot devices: Demo
  abstract,'' in \emph{Proceedings of the 14th ACM Conference on Embedded
  Network Sensor Systems CD-ROM}, 2016, pp. 288--289.

\bibitem{9606685}
V.~Niculescu, L.~Lamberti, F.~Conti, L.~Benini, and D.~Palossi, ``Improving
  autonomous nano-drones performance via automated end-to-end optimization and
  deployment of dnns,'' \emph{IEEE Journal on Emerging and Selected Topics in
  Circuits and Systems}, vol.~11, no.~4, pp. 548--562, 2021.

\end{thebibliography}
